# Reinforced Symbolic Learning with Logical Constraints for Predicting Turbine Blade Fatigue Life

Pei Li[1], Joo-Ho Choi[2], Dingyang Zhang[1], Shuyou Zhang[1], Yiming Zhang[1]*

[1] The State Key Lab of Fluid Power & Mechatronic Systems, Zhejiang University, Hangzhou 310058, PR China.
[2] School of Aerospace and Mechanical Engineering, Korea Aerospace University, Gyeonggi-do, 412-791, South Korea.

## Abstract

Accurate prediction of turbine blade fatigue life is essential for ensuring the safety and reliability of aircraft engines. A significant challenge in this domain is uncovering the intrinsic relationship between mechanical properties and fatigue life. This paper introduces Reinforced Symbolic Learning (RSL), a method that derives predictive formulas linking these properties to fatigue life. RSL incorporates logical constraints during symbolic optimization, ensuring that the generated formulas are both physically meaningful and interpretable. The optimization process is further enhanced using deep reinforcement learning, which efficiently guides the symbolic regression towards more accurate models. The proposed RSL method was evaluated on two turbine blade materials, GH4169 and TC4, to identify optimal fatigue life prediction models. When compared with six empirical formulas and five machine learning algorithms, RSL not only produces more interpretable formulas but also achieves superior or comparable predictive accuracy. Additionally, finite element simulations were conducted to assess mechanical properties at critical points on the blade, which were then used to predict fatigue life under various operating conditions.

## 1 Introduction

As a critical component of aircraft engines, the fatigue life of turbine blades is directly related to the safety and reliability of the aircraft. Aircraft engines typically operate under high temperature, high pressure and high speed, which can lead to various failure modes in turbine blades. Among these, fatigue damage caused by alternating loads represents a prevalent failure type observed in blade structures[1]. The inherent complexity of failure mechanisms, coupled with the microscopic scale of crack initiation, renders defect detection challenging. Consequently, the accurate and efficient prediction of turbine blade fatigue life has become a focal point of research within academic and industrial spheres. One of the most widely used approaches is to combine historical or real-time critical parameters measured by sensors on the blade, with failure cases to predict potential blade failures through time-series analysis[2]. In the absence of failure cases and data, researchers have initiated an effort to accurately delineate the intrinsic correlation between the mechanical properties of a material (e.g., tensile strength, ductility) and its fatigue characteristics, to develop fatigue life prediction models.

*Corresponding author.
E-mail address: yimingzhang@zju.edu.cn (Y. Zhang).

Several classic empirical formulas have been derived from traditional theoretical methods to establish the relationship between mechanical properties and fatigue life. Manson et al.[3]pioneered this effort by introducing the Four-Point Correlation Method and the Universal Slope Method, which preliminarily predict uniaxial fatigue life based on stress-life (S-N) curves and tensile test data within the framework of elastic deformation theory. Building upon these foundational methods, Muralidharan[4] incorporated tensile strength into the mechanical property parameters, and Ong et al.[5, 6] enhanced the Four-Point Correlation and Universal Slope Methods by employing multiple linear regression analysis. Roessle and Fatemi[7] identified a correlation between the hardness of steel and its fatigue life. They proposed a simplified method that relies solely on hardness and elastic modulus, effectively predicting strain-life curves for steel with hardness values ranging from 150 to 700 HB. While these approaches provided valuable insights, they primarily focused on uniaxial loading, whereas engineering components often encounter multiaxial loading conditions[8]. The Coffin-Manson equation utilized Von Mises stress to estimate elastic and plastic strain amplitudes in axial directions, accurately predicting the cycles to failure using an equivalent strain parameter. However, the maximum shear strain alone was insufficient to describe fatigue damage under uniaxial loading in multiaxial fatigue scenarios. This led to the introduction of a second damage control parameter, normal strain on the maximum shear plane, which underpins the development of the Brown-Miller model and the KBM model[9, 10]. To address the complexities of multiaxial fatigue, Fatemi and Socie[11] proposed the FS model, a multiaxial fatigue parameter based on shear strain that accounts for additional hardening effects. Under multiaxial fatigue loading, material cracks may initiate on the plane of maximum shear strain and propagate on the plane with the highest normal strain. The WHS model was subsequently established[12, 13], integrating maximum shear strain, normal stress, and strain on the maximum shear plane. Li et al.[14] proposed a nonlinear strength degradation model based on damage theory, which achieved higher prediction accuracy than SWT and FS models under different multiaxial loading. These empirical formulas are inherently reliant on extensive experimental data, which poses challenges in adapting models to complex operating conditions.

Data-driven machine learning algorithms provide new perspectives on fatigue life prediction, accurately capturing specific links between mechanical performance parameters and fatigue life without formulas, improving aircraft performance[15]. Zhu et al.[16] introduced Auto Gluon, an approach based on Random Forest (RF) and Support Vector Regression (SVR) algorithms, to predict the high-cycle fatigue life of titanium alloy TC17, achieving remarkable accuracy. K. Genel[17] explored the application of Artificial Neural Networks (ANN) for predicting strain-life fatigue properties by deploying four independent neural networks on tensile data from 73 types of steel. The prediction accuracy for the fatigue strength coefficient and the fatigue ductility (strain) coefficient reached 99% and 98%, respectively. He et al.[18] tested the performance of ANN, SVR, and RF algorithms in predicting the fatigue life of three different metal materials. Although the evaluated material properties varied across models, the optimal models demonstrated prediction results distributed within a twofold error margin. However, the performance of ML models heavily depends on sufficient datasets. To address the challenge of limited original datasets for additively manufactured Ti-6Al-4V samples, Jan Horňas et al.[19] proposed a novel approach combining dataset expansion methods—inverse transform sampling and multivariate radial basis function (RBF) interpolation—with a Gradient Boosting Regression (GBR) model. This method expanded the mechanical property parameters and significantly improved fatigue life prediction accuracy, achieving a coefficient of determination ($R^2$) of 0.953 on the test set and reducing the RMSE by 41.99%. Traditional multiaxial fatigue life prediction models are typically limited to specific materials and loading conditions, making them difficult to generalize. Yang et al.[20] proposed a multiaxial fatigue life prediction method that integrates Fully Connected Neural Networks (FCNN) and Long Short-Term Memory (LSTM) networks. This approach demonstrated superior predictive capabilities across various materials and complex loading conditions. Despite the powerful and flexible fitting capabilities of ML models, they often function as black-box models without providing transparency into the underlying decision-making processes. Nor can they explain the physical meaning or rationality of these parameters, which presents a challenge in elucidating the intrinsic link between material properties and fatigue life.

In practical engineering, there is a strong preference for concise, interpretable expressions that not only capture the relationships between data but also align with physical laws and established theories. Symbolic Regression (SR) is an evolutionary algorithm-based model that generates interpretable and generalizable expressions to describe the intrinsic relationships within data[21]. SR has demonstrated significant potential across various scientific and engineering domains, particularly in applications where understanding model behavior, adhering to physical principles, and making succinct predictions are paramount[22]. By optimizing both the structure and parameters of the expression models, SR produces explicit formulas that are computationally efficient and easily interpretable, offering a novel approach for predicting the fatigue life of critical components. Feng et al.[23] applied SR for feature engineering analysis and successfully predicted the corrosion fatigue life of T91 steel and 316L stainless steel (SS) used in fourth-generation advanced nuclear power plants. Their approach demonstrated a 22% reduction in RMSE compared to Artificial Neural Networks (ANN), preliminarily confirming the efficiency of SR in fatigue life prediction. Similarly, Cao et al.[24] proposed two SR algorithms based on Genetic Programming—GP-SR and MPEA-SR, which derived general formulas accurately linking four material performance parameters with hardness and fatigue performance. Zhou[25] proposed a SR algorithm guided by domain knowledge to develop a predictive model for material fatigue crack growth (FCG) rates. Compared to three other traditional semi-empirical FCG models, this SR-based model accounted for the effects of compressive loads and provided more accurate predictions under different load ratios. Zhao et al.[26] developed a model using unbiased SR techniques to predict macroscopic hardness and fracture toughness, with shear and bulk modulus as inputs. The model's predictive performance was comparable to that of a Graph Convolutional Neural Network, and it efficiently screened for potential superhard materials with desirable fracture toughness.

These studies underscore the advantages of SR in overcoming the limitations of empirical formulas, such as insufficient prediction accuracy and a heavy reliance on prior theoretical knowledge. Additionally, SR offers greater interpretability and simpler structure compared to traditional ML models. Nonetheless, further research is needed to enhance the theoretical foundation and practical applications of SR in the context of fatigue life prediction. This study proposes a novel framework RSL, combining symbolic regression with reinforcement learning, to explore the intrinsic relationship between fundamental mechanical properties and the fatigue life of materials. Using GH4169 and TC4 materials from aircraft engine turbine blades as case studies, our RSL method not only discovered more accurate predictive formulas but also precisely predicted the fatigue life of materials under various operating conditions. This approach offers technical support and introduces new perspectives for advancing research on the fatigue life of critical components in aircraft engines. The main contributions of this paper are as follows：

(1) Development of the RSL Framework: We propose a comprehensive Reinforced Symbolic Learning (RSL) framework that combines symbolic regression with deep reinforcement learning, designed to predict fatigue life with high interpretability and physical relevance.

(2) Introduction of Logical Constraints: The RSL framework incorporates logical constraints within the symbolic regression process, including restrictions on the length of expressions, the unity of physical units, and the nesting of functions, etc., to ensure that the derived formulas are structurally valid and aligned with physical laws, significantly improving the efficiency of the search process.

(3) Enhancements with Deep Reinforcement Learning: We enhance the symbolic regression process with deep reinforcement learning in the form of RNN, which guides the optimization process, enabling the discovery of more accurate and generalizable predictive models.

(4) Application to Turbine Blade Materials: The RSL method is validated using two widely used turbine blade materials, GH4169 and TC4, effectively predicting fatigue life cycles under various conditions and demonstrating superior performance compared to empirical formulas and machine learning models.

The organization of the paper is as follows: Section 2 overviews the technical background, including a review of existing methods for fatigue life prediction and an introduction to symbolic regression. The core methodology is then presented in Section 3, detailing the Reinforced Symbolic Learning (RSL) framework and its components, such as logical constraints and deep reinforcement learning. Next, Section 4 details the experimental setup and numerical results, evaluating the application of RSL to GH4169 and TC4 materials and comparing its performance with alternative methods. Finally, Section 5 concludes the findings and suggestions for future research directions.

# 2 Technical background

Accurate fatigue life prediction of turbine blades is essential for aircraft engine reliability. Traditional empirical methods, while widely used, often lack precision under complex conditions. Machine learning has improved predictive accuracy but often at the cost of interpretability. This section reviews the traditional methods to characterize the mechanical properties in Section 2.1 and the basis of symbolic regression in Section 2.2.

## 2.1 Turbine blade life prediction

The prediction of fatigue life for turbine blades has increasingly focused on accurately capturing the intrinsic relationship between material mechanical properties and fatigue life. Empirical formulas, grounded in theoretical principles, remain one of the most widely used methods in engineering for this purpose. Below are some representative empirical formulas for predicting fatigue life.

**(1) Coffin-Manson criterion**

Given that localized plastic deformation during persistent slip in materials leads to the initiation of fatigue cracks, and that the slip bands align with the direction of maximum shear strain, researchers have employed the Coffin-Manson equation based on maximum shear strain to predict the multiaxial fatigue life of materials[27].

$$\gamma_{a,\max} = \frac{\tau_f^{'}}{G}(2N_f)^{b_0} + \gamma_f^{'}(2N_f)^{c_0} \tag{1}$$

The term $\gamma_{a,max}$ represents the maximum shear strain amplitude.

**(2) BM and KBM criterion**

Kandil and Miller put forth the idea that cyclic shear strain has the effect of promoting the initial formation of cracks, while normal strain is responsible for driving the subsequent propagation of these cracks. They proposed that the critical plane is determined by a combination of normal and shear strains. Based on this premise, they developed the strain-based KBM fatigue criterion, which was derived from the BM criterion[9, 10].

$$\frac{\Delta\gamma_{\max}}{2} + s_0\Delta\varepsilon_n = \left[1+v_e+s_0(1-v_e)\right]\frac{\sigma_f^{'}}{E}(2N_f)^b + \left[1+v_p+s_0(1-v_p)\right]\varepsilon_f^{'}(2N_f)^c \tag{2}$$

$$s_0 = \frac{\frac{\tau_f^{'}}{G}(2N_f)^{b_1} + \gamma_f^{'}(2N_f)^{c_1} - (1+v_e)\frac{\varepsilon_f^{'}}{E}(2N_f)^b - (1+v_p)\varepsilon_f^{'}(2N_f)^c}{(1-v_e)\frac{\sigma_f^{'}}{E}(2N_f)^b + (1-v_p)\varepsilon_f^{'}(2N_f)^c} \tag{3}$$

where $\Delta\gamma_{max}/2$ represents the maximum shear strain amplitude, $\Delta\varepsilon_n$ is the range of normal strain on the maximum shear plane, and $s_0$ is a material constant obtained by fitting uniaxial and torsional fatigue data. In instances where data is unavailable, certain fatigue parameters can be approximated. For instance, $\tau_f' \approx \sigma_f'/\sqrt{3}$, $\gamma_f' \approx \sqrt{3}\varepsilon_f'$, $b \approx b_1$, $c \approx c_1$.

(3) **FS criterion**

The KBM criterion is limited to consideration of strain components. Building on the KBM model, Fatemi and Socie[11] introduced a new model (FS model) by replacing the normal strain with the maximum normal stress on the critical plane.

$$\frac{\Delta\gamma_{\max}}{2}\left(1+k\frac{\sigma_{n,\max}}{\sigma_y}\right)=\frac{\tau_f'}{G}(2N_f)^{b_0}+\gamma_f'(2N_f)^{c_0} \tag{4}$$

$$k=\left[\frac{\frac{\tau_f'}{G}(2N_f)^{b_1}+\gamma_f'(2N_f)^{c_1}}{(1+v_e)\frac{\sigma_f'}{E}(2N_f)^{b}+(1+v_p)\varepsilon_f'(2N_f)^{c}}-1\right]\frac{2\sigma_y}{\sigma_f'(2N_f)^{b}} \tag{5}$$

In this model, the material constant $k$ is obtained by fitting uniaxial fatigue data with pure torsion data, thereby representing the material's sensitivity to normal stress. $\sigma_{n,max}$ is the maximum normal stress amplitude.

(4) **WHS and MWHS criterion**

The FS model demonstrates a high degree of accuracy in the prediction of shear crack modes. However, in multiaxial fatigue tests, mixed crack modes are frequently observed. In order to address this issue, researchers developed the WHS model[12, 13] by incorporating maximum shear strain, normal stress, and strain on the maximum shear plane.

$$\frac{\Delta\gamma_{\max}}{2}+k\left(\frac{\sigma_{n,\max}\Delta\varepsilon_n}{E}\right)^{0.5}=\frac{\tau_f'}{G}(2N_f)^{b_0}+\gamma_f'(2N_f)^{c_0} \tag{6}$$

It is noteworthy that in WHS model, the material parameter $k$ is not constant but varies with the fatigue life of the material. Based on WHS criterion, minor alterations were made, resulting in the creation of the MWHS model, which is outlined below:

$$\frac{\Delta\gamma_{\max}}{2}+\frac{1}{2}\left(1+\frac{\sigma_{n,\max}}{\sigma_y}\right)\left(\frac{\sigma_{n,\max}\Delta\varepsilon_n}{E}\right)^{0.5}=\frac{\tau_f'}{G}(2N_f)^{b_0}+\gamma_f'(2N_f)^{c_0} \tag{7}$$

The accuracy of empirical formulas is frequently constrained by the sufficiency and representativeness of experimental data. Although these formulas demonstrate strong generalizability, their predictive accuracy for specific materials utilized in engine blades remains a subject for improvement. In light of the fact that machine learning (ML) models are capable of attaining high levels of precision in data fitting, our study explores several representative data-driven ML algorithms, including Support Vector Regression (SVR), Random Forest (RF), Long Short-Term Memory (LSTM) networks, XGBoost, and Gradient Boosting. Each algorithm has distinctive capabilities in addressing non-linear relationships and discerning intricate patterns, which contributes to enhanced predictive accuracy.

Despite the ability of ML models to precisely fit data and achieve high predictive accuracy, they frequently encounter difficulties in elucidating the underlying physical relationships within data or ensuring that predictions adhere to reasonable physical constraints. In the case of critical components such as blades, it is currently preferable to establish explicit relationships based on theoretical foundations to predict fatigue life.

## 2.2 Fundamentals of symbolic regression

Symbolic regression (SR) is particularly adept at providing explanations for extensive experimental and observational results, generating compact, interpretable, and generalizable expression. In particular, when a dataset $(X, Y)$ is provided, where each point satisfies $X_i \epsilon R^n$, $Y_i \epsilon R$, SR is able to identify the optimal function $f$ that maps $R^n \rightarrow R$, to a concise mathematical expression that is capable of explaining or providing effective scientific laws.

Despite the considerable potential of SR in the discovery of physical laws and expressions, there are significant challenges in its practical application. The essence of the expression generation process lies in the selection of appropriate nodes from a symbolic library to form sequences. This process requires the rational selection of function symbols (e.g., +, -, *, /, log), input variables (e.g., $x_1, x_2$), and parameters, as well as the determination of specific parameter values. The symbolic space is discrete and grows exponentially with the length of expression. A brute-force search would encounter issues such as an enormous search space and the inability to construct structurally sound expressions[28]. Traditional SR frequently employs evolutionary algorithms, such as selection, crossover, and mutation in genetic programming, to iteratively improve the fitness function. However, these methods are highly sensitive to hyperparameters and exhibit slow convergence in large-scale problems. With the continuous advancement of deep learning, researchers are attempting to link deep learning with SR by using neural networks to search the space of symbolic nodes.

Incorporating the physical context, the majority of the variables introduced in model possess physical units. Therefore, it is essential to ensure that the expression has a reasonable structure and to maintain the balance of physical units on both sides of the equation. To address this issue, some SR algorithms, such as PhySO, attempt to introduce a system that balances physical units and dimensions[21]. Nevertheless, this methodology necessitates the development of a novel dimensional system, specifically tailored to the given application domain. This often involves intricate derivations of units, particularly in fields such as astrophysics. In contrast, the AI Feynman algorithm[29] and the method proposed by Matchev et al.[30] employ dimensional analysis to identify dimensionless combinations of input variables, converting symbolic variables into dimensionless structures and reducing the number of independent input variables. Sebastian[31] introduced a novel variational autoencoder for SR, demonstrating that the application of logical constraints can significantly enhance the performance of the algorithm. In particular, it was observed that filtering expressions through logical constraints resulted in a reduction in error with 2,000 training expressions, in comparison to 12,000 unconstrained samples. In light of the algorithm's efficiency and the logical framework's simplicity, we employ partial dimensionless processing and logical rule constraints. By preventing the generation of structures that violate logical constraints, the symbolic search space has been reduced.

In this paper, we represent expressions using a binary tree structure, analogous to those employed in genetic programming (GP), and rationality constraint rules are applied to guarantee that the generated expressions are both reasonable and valid. Furthermore, physical logic constraints are imposed to preserve the equilibrium of physical units and guarantee that the generated expressions align with the actual physical significance. We also incorporate a risk-seeking policy gradient approach, rigorously training our neural network to strictly follow these rules under the guidance of reinforcement learning.

# 3 The proposed Reinforced Symbolic Learning

This section presents the Reinforced Symbolic Learning (RSL) framework developed for predicting turbine blade fatigue life. Section 3.1 introduces the RSL framework to derive accurate, interpretable predictive models. Section 3.2 details the integration of logical constraints, and Section 3.3 discusses the optimization process guided by deep reinforcement learning.

## 3.1 Overall framework

In the Reinforced Symbolic Learning (RSL) framework, expressions generated through symbolic regression (SR) are represented as binary trees as shown in Fig. 1. Each node in the tree corresponds to a symbol from a predefined library, which includes function operators, input variables, and constants. Unary operators, such as *sin* and *log*, perform operations on a single parameter and have one child node in the binary tree, while binary operators, such as addition, subtraction, division, and multiplication, operate on two parameters and have two child nodes. Input variables represent the material's mechanical properties, and both these variables and constants serve as terminal nodes, meaning they do not have child nodes. Once the required child nodes for each function node are assigned, the tree structure is complete, and the corresponding expression is generated. The predictive accuracy of these expressions is evaluated using metrics such as root mean square error (RMSE) by comparing the predicted values with the actual dataset.

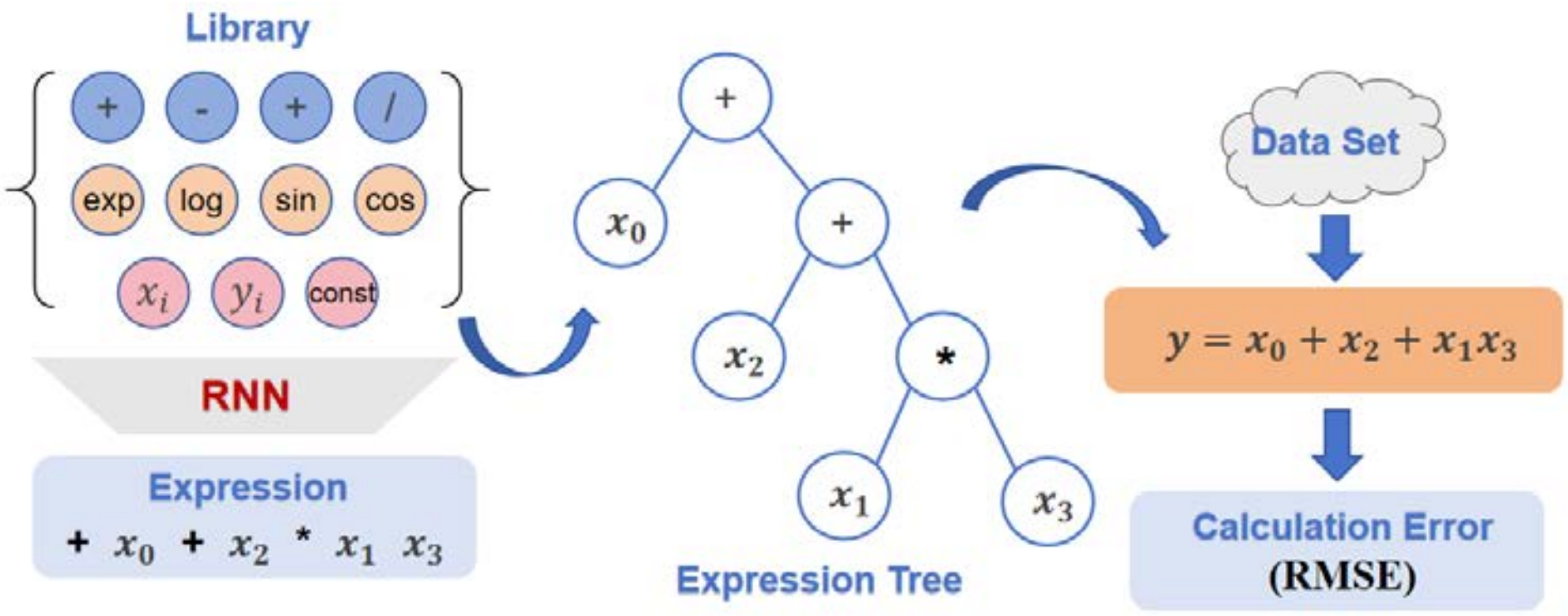

**Figure 1. The expression structure framework in RSL.**

When the expression tree is unfolded into a one-dimensional sequence, each node is generated by evaluating the probabilities of potential nodes using a Recurrent Neural Network (RNN). As each new node is added, it creates a time-dependent state stored in memory. To predict the next node, the RNN takes all accumulated states as input. Guided by a policy function, the RNN produces a probability distribution over possible symbols, selecting the one with the highest probability for the next node. The sequence, along with the updated count of pending nodes and the relevant logical constraints, is then fed back into the RNN to generate subsequent nodes. This process continues iteratively until the entire expression structure is produced, as shown in Fig. 2.

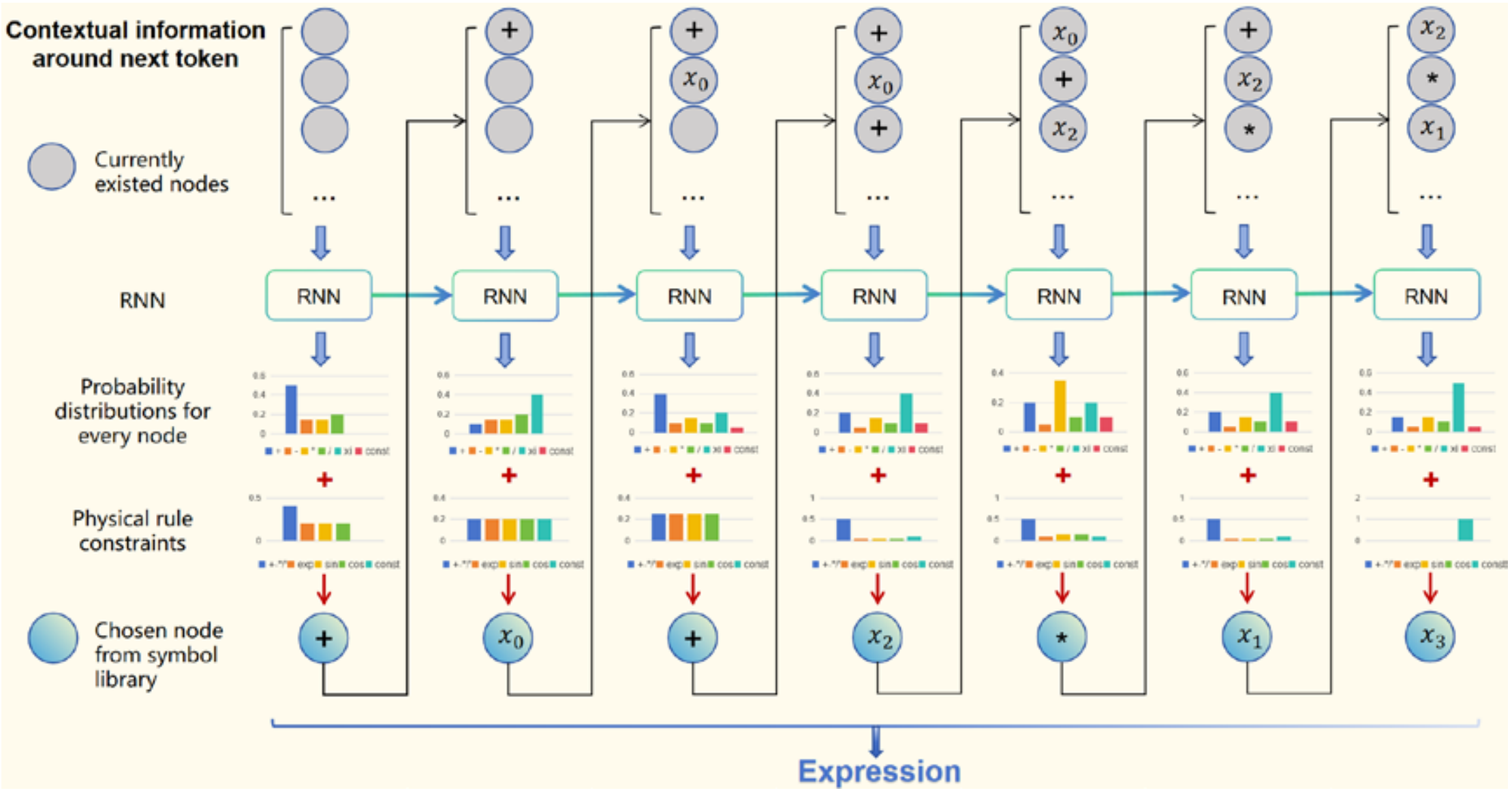

**Figure 2. Schematic of the RNN-guided selection process for each node in expression sequence.**

As the sequence progresses, the current expression, the number of pending nodes, and the necessary logical constraints are updated and used as input for the RNN to generate subsequent nodes. This RNN-guided process repeats until the entire expression structure is generated. Given the vast search space for symbols, many unfeasible structures can arise. To address this, the RSL algorithm incorporates constraint rules that significantly reduce the search space and ensure the generated expressions maintain physical relevance.

## 3.2 Formulation and logical constraints of SR expressions

In symbolic regression (SR), the structural and logical validity of generated expressions is crucial. To enhance the simplicity and rigor of the expression structures, this study introduces a series of constraint mechanisms as shown in Fig. 3. These mechanisms not only control the length of the expressions and ensure the consistency of physical units, but also impose rigorous constraints on constant calculations and function nesting.

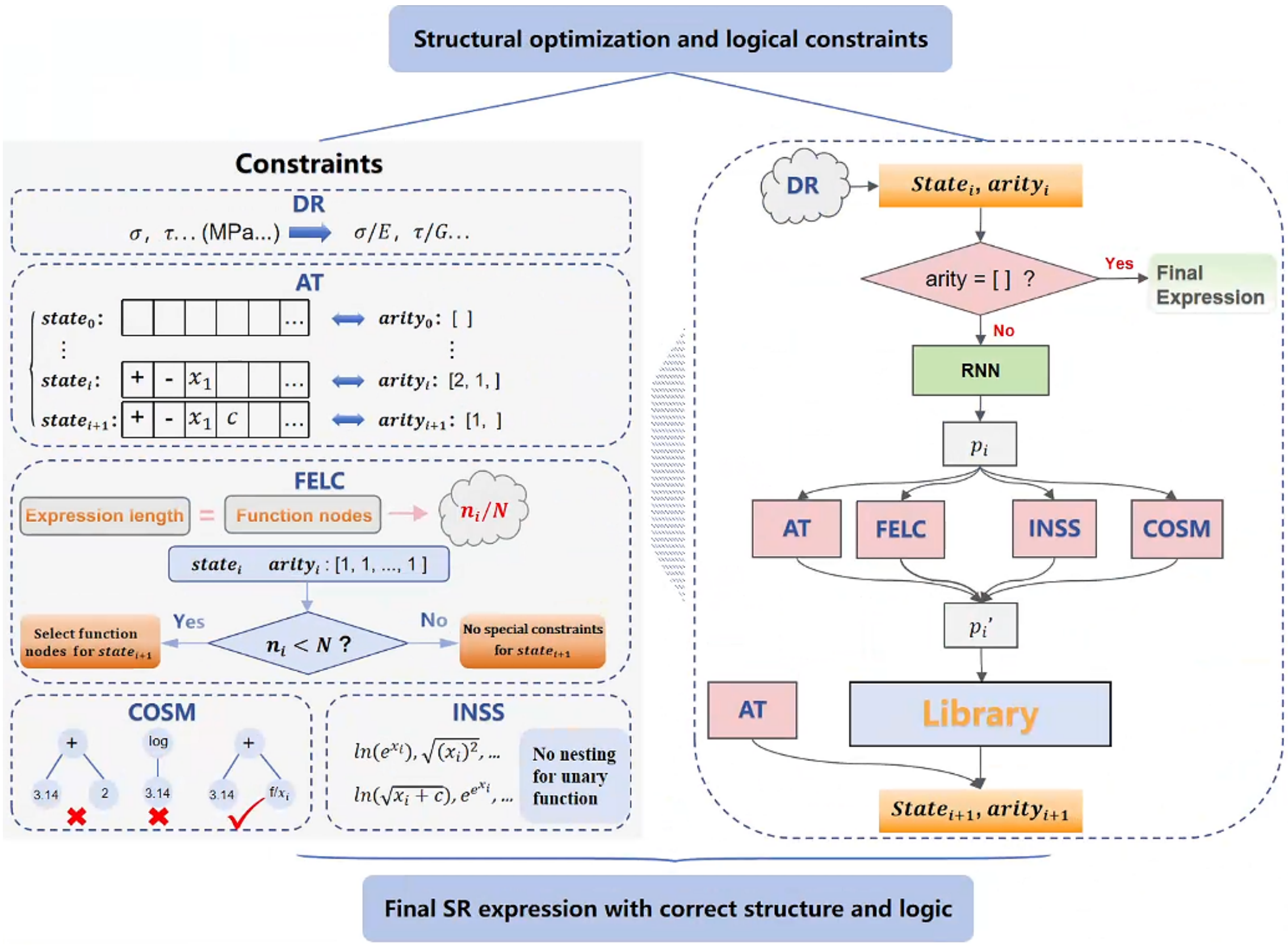

**Figure 3. Flowchart for imposing structural simplifications and logical constraints on expressions.**

**Dimension Removal (DR):** Given that the input mechanical parameters frequently possess physical units (e.g., MPa, GPa), we incorporated partial dimensionless constraint rules to guarantee the consistency of physical units within the expressions. This rule entails the transformation of variables with physical units into dimensionless combinations that retain their physical significance.

**Arity Tracking (AT):** During the construction of the expression tree, an initial empty "arity" array is generated, which is updated to reflect the number of child nodes required by each function node. This allows for the monitoring of the progress of the expression's construction. Each $state_i$ is associated with a corresponding $arity_i$. Specifically, when a new function node is introduced, the requisite number of child nodes for that function is appended to the end of the arity array. Upon the addition of a constant or variable, the final value in the arity array is decremented by one. Should this value reach zero, it is removed, and the process resumes with the preceding value. The absence of any values in the arity array signifies the completion of the expression's construction.

**Function Enforcement for Length Constraint (FELC):** To regulate the length of expression and guarantee that it matches the complexity of the data, our RSL algorithm constraints the number of function nodes within the expression. Additionally, to enhance the diversity of generated expressions, the algorithm alternates between generating function nodes and constants. However, this approach may result in the expression tree being fully constructed (i.e., the arity array is empty) before reaching the desired expression length. To circumvent such complications, a manual constraint is imposed on RSL. If the expression has not yet reached the predetermined length and the arity array is comprised solely of 1s, a function node is forcibly selected as the subsequent node.

**Constant Operation Suppression Mechanism (COSM):** In the majority of cases, the result of a constant calculation can be replaced by a single constant. To keep the expression as concise as possible, RSL forbids function nodes from having only constants as child nodes (e.g., log(3.14) is not permitted, but the formation like $\sqrt{3.14 + x}$, is permitted. Our research has shown that the number of constant nodes significantly impacts the structure of expression. Without constraints, the algorithm tends to explore constant nodes excessively, substituting them for variable parameters. Accordingly, a threshold is imposed on the number of constant nodes, which can be adjusted based on the desired expression length and the complexity of the dataset.

**Irrational Nesting Suppression Strategy (INSS):** In practical physical formulas, nested unary function operations are exceedingly uncommon and frequently devoid of any tangible physical significance. Furthermore, exponential operations are highly computationally intensive, and their nesting can result in a significant increase in computational range, which can negatively impact performance. And to avoid redundant and meaningless inverse unary operations such as $\ln{(e^x)}$ and $e^{\ln x}$, we prohibit the nesting of other unary function nodes within a unary function node.

To enhance efficiency and minimize the generation of expressions that do not meet the constraints, prior rule constraints are applied during the selection process for each node (i.e. during the update of $state_i$). Consequently, in addition to the probabilities $p_i$ generated by the RNN for each node in the library, we incorporate an additional layer of logical constraint probability, assigning a probability of -inf to nodes that violate the constraints. The combination of the RNN's probability distribution with that of the logical constraint probability enables the selection of the most appropriate new node, thus ensuring that the constructed expression is both structurally sound and physically meaningful.

## 3.3 RNN-based Reinforcement Learning for expression generation

In this paper, the selection of nodes within the training network is guided by a reinforcement learning strategy. The RNN serves as the core of this process, generating a probabilistic distribution over the symbols in the library, which acts as the policy within the reinforcement learning framework. The RNN produces a batch of expressions and reinforces behaviors associated with high reward values, guiding subsequent batches toward better results. This iterative process continues until the network consistently selects symbolic nodes that are well-suited to the data.

In the formulation of reinforcement learning algorithms, the selection of the next action based on the current expression state is determined by the policy function $\pi_\theta$. The policy function $\pi_\theta(a_t|s_t)$ represents the probability of selecting action $a_t$ given the current state $s_t$, and is expressed as follows:

$$\pi_\theta(a_t \mid s_t) = P(a_t \mid s_t; \theta) \tag{8}$$

where $s_t$ represents the state during the expression generation process, which corresponds to the current symbol sequence; $a_t$ is the next node selected from the symbol library; and $\theta$ denotes the weight parameters of the RNN network, which are optimized through training. Following the policy function $\pi_\theta$ to select a new node and add it to the expression sequence in the current state $s_t$ yields a corresponding action reward. The expected cumulative reward can be expressed by the state value function $V^{\pi}(s)$, and the Bellman Expectation Equation describes the expected value of $V^{\pi}(s)$ under a given policy as follows:

$$V^{\pi}(s) = E_{a\sim\pi}\left[R(s,a) + \gamma V^{\pi}(s')\right] \tag{9}$$

where $s'$ represents the next state after executing action $a$, and the impact of the expected cumulative reward $V^{\pi}(s')$ on the current reward is measured by the discount factor $\gamma$. $R(s,a)$ denotes the immediate reward function associated with taking action $a$ in state $s$. Accordingly, based on the state value parameters, the state-action value function $Q^{\pi}(s,a)$ represents the expected cumulative reward after choosing action $a$ in state $s$ and is expressed as follows:

$$Q^{\pi}(s,a) = R(s,a) + \gamma E_{s'\sim P}[V^{\pi}(s')] \tag{10}$$

The Q function can be used to evaluate the quality of each generated symbol (action) and guide policy optimization. By reinforcing behaviors associated with higher reward values during the training of the RNN, the network is encouraged to generate expression structures that better fit the data based on the sequence environment. To improve the accuracy of the policy function $\pi_\theta$ in selecting nodes, the parameters $\theta$ in $\pi_\theta$ must be continually optimized. This optimization is achieved through policy gradients, where the policy is updated by maximizing the expected reward $J(\theta)$.

$$\nabla_\theta J(\theta) = E_{s\sim\pi_\theta, a\sim\pi_\theta}\left[\nabla_\theta \log \pi_\theta(a \mid s) R(s,a)\right] \tag{11}$$

where $\log \pi_\theta(a|s)$ represents the logarithm of the policy function's probability, and its gradient is used to adjust the parameters $\theta$ to increase the probability of selecting high-reward actions. The optimization objective in reinforcement learning is to maximize the cumulative expected reward, which is expressed in the policy gradient as:

$$J(\theta) = E_{\pi_\theta}\left[\sum_{t=0}^{T} \gamma^t R_t\right] \tag{12}$$

In this context, $T$ represents the length of the time series, whereas $R_t$ is the reward at time step *t*. The accuracy of the expression's prediction is evaluated through the RMSE between the predicted and actual values. To maximize the reward obtained by the expression, we define *R* as the negative value of RMSE. This approach encourages the generation of expressions that result in lower RMSE values.

When training the network parameters, we select the subset of expressions with the lowest RMSE and the best fitting performance from each batch, forming a group of elite expressions. Subsequently, these elite expression structures are subjected to further reinforcement. Sebastian et al.[31] demonstrated that applying this risky gradient search strategy to the optimization of expression structures significantly improved the effectiveness of symbolic regression work. Additionally, we balance exploration of new strategies and exploitation of the current best strategy to ensure diversity in expression structures. During the initial stages of expression generation, the neural network's node selection may not be highly accurate, resulting in a lower overall quality of the generated expressions. To address this issue, the number of expressions in the initial batches is increased in order to explore a wider variety of structures, while a more stringent evaluation threshold is applied to elite expressions. Ultimately, we choose a batch size of 1,300, selecting the top 4% as elite expressions, equating to 52 expressions. In the initial two batches, the size of the batch was increased to 16,900, with the top 2.5% selected as elite expressions, resulting in 423 selected expressions.

From this, it is clear that the generation of nodes relies on the RNN network observing all previous states and selecting nodes from the library based on the combined sequence of past states $s$ and the policy function.

Unfortunately, when generating the first node of an expression, there is no prior state $s$ for RNN to observe, making it impossible for reinforcement learning to learn a strategy for generating the initial node based on the environment. To address this issue, Wassim Tenachi et al.[21] filled the first node with a blank value and considered the sequence starting from the second node as the observation sequence. Our approach entails a statistical analysis of the type and occurrence probability of the first node in the elite expressions from the previous batch and use this probability to select the first node in the current batch. This enables the generation strategy for all nodes to be learned, thereby facilitating the production of accurate expressions that align with the real dataset.

Free constants within the expression can simulate unknown physical parameters while also enhancing the expression's fitting accuracy to real data. During the selection of nodes, a constant placeholder is generated, but the specific value of the constant remains unknown. We use the LBFGS optimizer to adjust the constant nodes, thereby identifying the optimal constant values that minimize the fitting error of the expression. The LBFGS gradient descent formula is as follows:

$$\theta_{\text{optimal}} = \arg\min_{\theta} RMSE\left(y_{true}, y_{pred}\right) \tag{13}$$

The utilization of reinforcement learning to direct the selection of each node enables the generation of expressions that are accurately aligned with the dataset. Given the generalizability of physical expressions, particularly across different conditions for the same material, our objective is to identify a universal expression that maintains a consistent structure. Consequently, following the identification of the expression that yields the optimal fit, its sequence structure is preserved, and the optimal constant values are explored using experimental datasets from different working conditions. This approach guarantees that the expression will remain generalizable across analogous materials.

# 4 Numerical evaluations

This section presents the numerical evaluations of the Reinforced Symbolic Learning (RSL) framework applied to turbine blade fatigue life prediction. Section 4.1 outlines the experimental setups, including the methods used for comparison. Section 4.2 assesses the predictive performance of RSL through experiments on GH4169 and TC4 materials, comparing the results with traditional empirical formulas and machine learning models. Section 4.3 integrates finite element simulations with RSL for turbine blade fatigue life prediction.

## 4.1 Experimental setup

In order to validate the accuracy of the RSL model in predicting the fatigue life of turbine blades in aircraft engines, experimental data from commonly used blade casting materials were employed: the nickel-based superalloy GH4169 and the titanium alloy TC4, under different loading paths[12, 27, 32]. Due to the constraints of the available experimental data, certain shear fatigue characteristic parameters were estimated using empirical formulas. The final fatigue performance parameters of the materials are attached in Tables A1- A3 in the Appendix. Tubular specimens with an outer diameter of 14 mm/17 mm and a gauge length of 32 mm were employed for multiaxial fatigue testing, as detailed in references[12, 32]. The tests were conducted under strain-controlled conditions, with identical specimen dimensions maintained across all testing scenarios. The fatigue test data under multiaxial symmetric loading are provided in Table A4, located in the Appendix.

Prior to commencing the RSL algorithm, it is necessary to define a number of crucial hyperparameters as initializations. The specific hyperparameter settings are shown in Table 1.

**Table 1. Initial hyperparameter settings for the RSL algorithm.**

| Parameters | $N_{size}$ | $N_{epoch}$ | $n_{group}$ | $p_1$ | $p_2$ | $l$ | $N_{const}$ |
|---|---|---|---|---|---|---|---|
| **Values** | 1300 | 32 | 2 | 0.025 | 0.04 | 10 | 5 |

$N_{size}$ represents the number of expressions generated in each batch. $N_{epoch}$ denotes the number of expression generation epochs, and $n_{group}$ indicates the number of epochs during which expression augmentation is performed. $p_1$ represents the proportion of elite expressions selected in the augmented batches, while $p_2$ denotes the proportion of elite expressions selected in subsequent batches. As discussed in Section 3.3, the number of expressions was increased from 1,300 to 16,900 in the initial two epochs, so the selection proportion of elite expressions in the initial batches is lower than in the later batches. $l$ represents the length of the expression, and $N_{const}$ denotes the maximum number of constants allowed in an expression. These hyperparameters provide structural constraints on the expressions, initially limiting their complexity. They can be adjusted later based on the complexity and scale of the dataset to improve the expression's fitting accuracy to real data. The setup details of the experiment are shown in Table 2.

**Table 2. Details of experimental setups.**

| Number | Factor | Setup details |
|---|---|---|
| 1 | Date set | GH4169 at 25℃(Data 1), TC4 at 25℃(Data 2), GH4169 at 650℃(Data 3) |
| 2 | Number of data | 18×3 |
| 3 | Alternative-empirical formula | Coffin-Manson, BM and KBM, FS, WHS, MWHS |
| 4 | Alternative- ML algorithm | LSTM, SVR, RF, Gradient Boosting, XGBoost |
| 5 | Accuracy metrics | RMSE, $R^2$ |

Following the generation of each expression, it must be evaluated in order to ascertain its fitting accuracy when applied to the actual data set. A subset of the most appropriate expressions is then selected from the batch to train RNN network. In this project, the evaluation criteria employed are the root mean square error (RMSE) and the coefficient of determination ($R^2$). RMSE is a measure of the discrepancy between the observed and true values, with lower values indicating greater accuracy in the predictions. The $R^2$ ranges from [0,1], with a value closer to 0 indicating a poor fit of the model to the data, and a value closer to 1 indicating a better fit. To demonstrate the accuracy and reliability of RSL in predicting fatigue life, a comparison is made between the predictive performance of RSL and six typical fatigue life prediction empirical formulas and five commonly used ML algorithms, as discussed in Section 2.1.

## 4.2 Modeling of the material properties

**(1) Data processing**

Given that the parameters involved in the engine blade dataset often include physical units, it is essential to maintain the balance of physical units on both sides of the expression equation. In accordance with the Dimension Removal constraints discussed in Section 3.2, we substitute axial stress $\sigma_a$ with $\sigma_a/E$ and shear stress $\tau_a$ with $\tau_a/G$, which are not only common in empirical formulas but also carries specific physical significance. Young's modulus $E$ is typically used to measure a material's ability to deform under axial stress, defining the linear relationship between stress and strain within the material. On a stress-strain curve, $E$ corresponds to the slope, meaning that $\sigma_a/E$ physically represents axial strain. Similarly, $\tau_a/G$ represents shear strain. In summary, the input parameters of the RSL model include $\varepsilon_a$, $\gamma_a$, $\sigma_a/E$ and $\tau_a/G$, while the output parameter corresponds to the fatigue life cycle count $N_f$ associated with the mechanical properties.

**(2) Modeling with RSL**

Following the hyperparameters delineated in Table 1 and setup details in Table 2, the preprocessed dataset was fed into RSL algorithm for the purpose of identifying well-fitted expressions. In addition to evaluating the fitting performance, consideration was given to the frequency of similar structural patterns and the generalizability of the expressions. For each dataset, three representative high-performing expressions were selected based on a comprehensive assessment, as detailed in Table 3. The expressions are ordered by RMSE in descending order, thereby reflecting the quality of the fit.

**Table 3. Representative high-performing life prediction formulas identified by RSL, along with the RMSE and $R^2$ values comparing predicted and actual outcomes.**

| Dataset | Formula Expression | RMSE | $R^2$ |
|---|---|---|---|
| GH4169 at 25℃ | $\ln(N_f) = 3.148 + \dfrac{7.171(\varepsilon_a \frac{\tau_a}{G} - 0.785\frac{\tau_a}{G} + 2.74)}{(\varepsilon_a + \gamma_a)(\varepsilon_a \frac{\tau_a}{G} - 0.785\frac{\tau_a}{G} + 2.74) - 0.003}$ | $9.8525 \times 10^2$ | 0.98766 |
| | $\ln(N_f) = 5.27 + \dfrac{\gamma_a + 11.443}{\varepsilon_a \frac{\tau_a}{G}} - \dfrac{0.048}{\varepsilon_a^2 \frac{\tau_a}{G}(\gamma_a + \frac{\tau_a}{G} - 6.81)}$ | $1.0067 \times 10^3$ | 0.98767 |
| | $\ln(N_f) = 5.259 + \dfrac{\gamma_a + 11.485}{\varepsilon_a \frac{\tau_a}{G}} - \dfrac{0.051}{\varepsilon_a^2 \frac{\tau_a}{G}(\varepsilon_a + \frac{\tau_a}{G} - 6.527)}$ | $1.0070 \times 10^3$ | 0.98766 |
| TC4 at 25℃ | $\ln(N_f) = 5.918 - \dfrac{6.173}{\gamma_a} + \dfrac{5.178}{\varepsilon_a \gamma_a} - \dfrac{2.827G}{(\varepsilon_a \gamma_a)^2 \tau_a}$ | $4.5838 \times 10^3$ | 0.91285 |
| | $N_f = \dfrac{-181.472\frac{\tau_a}{G}}{\varepsilon_a(-1.98\varepsilon_a \frac{\tau_a}{G} + 10.701\varepsilon_a + \frac{\tau_a}{G} - 5.807)}$ | $4.4164 \times 10^3$ | 0.91911 |
| | $\ln(N_f) = \dfrac{0.0029\varepsilon_a}{7.126 - \frac{\tau_a}{G}} + 0.0029\dfrac{\sigma_a}{E} + 4.24559 + \dfrac{2.2477}{\varepsilon_a}$ | $4.2178 \times 10^3$ | 0.92622 |
| GH4169 at 650℃ | $\ln(N_f) = 5.179 + \varepsilon_a((1.344 - \gamma_a)(\gamma_a - 7.092\frac{\varepsilon_a \tau_a}{G} + 4.117\frac{\tau_a}{G}) + 0.417)$ | $1.4452 \times 10^2$ | 0.99040 |
| | $\ln(N_f) = 4.968 + \varepsilon_a(\frac{\tau_a}{G}(1.391 - \gamma_a)(\varepsilon_a^2 - 7.85\varepsilon_a + 4.207) + 1.266)$ | $1.4531 \times 10^2$ | 0.99029 |
| | $\ln(N_f) = 0.302 + \dfrac{11.625(\frac{\varepsilon_a \tau_a}{G} + 297.975\frac{\tau_a}{G} + 185.977)}{(\varepsilon_a + \gamma_a)(\frac{\varepsilon_a \tau_a}{G} + 297.975\frac{\tau_a}{G} + 185.977) + 1202.181}$ | $2.9413 \times 10^2$ | 0.96022 |

Upon analyzing the fatigue life prediction formulas corresponding to these two materials, it becomes evident that the formula structures for GH4169 at 25°C and at 650°C are significantly more similar than those for TC4, and a highly similar formula structure may be identified. In contrast, the fatigue life prediction formulas for TC4 display a slightly different structure. We hypothesize that for the same type of material under different operating conditions, the RSL algorithm can discover a structurally generalized expression with different constant values, capturing the relationship between the material's mechanical properties and its fatigue life. While the expressions between different materials are inherently different in structure.

To validate this hypothesis, the fatigue test dataset of GH4169 at 25°C was used as input to the RSL algorithm to identify the optimal expression. The structure of this optimal expression was then retained, and the algorithm was used to find the best-fitting expressions for the GH4169 dataset at 650°C and the TC4 dataset at 25°C. The optimal expressions identified by RSL are presented in Table 4.

**Table 4. Optimal generalized formulas identified by RSL and the RMSE and $R^2$ values comparing predicted and actual outcomes.**

| Dataset | Best Expression | RMSE | $R^2$ |
|---|---|---|---|
| GH4169 at 25℃ | $\ln(N_f)=3.148+\dfrac{7.171(\varepsilon_a\frac{\tau_a}{G}-0.785\frac{\tau_a}{G}+2.74)}{(\varepsilon_a+\gamma_a)(\varepsilon_a\frac{\tau_a}{G}-0.785\frac{\tau_a}{G}+2.74)-0.003}$ | $9.8525\times10^2$ | 0.98766 |
| TC4 at 25℃ | $\ln(N_f)=-17.544+\dfrac{56.092(\varepsilon_a\frac{\tau_a}{G}+196.205\frac{\tau_a}{G}-45.141)}{(\varepsilon_a+\gamma_a)(\varepsilon_a\frac{\tau_a}{G}+196.205\frac{\tau_a}{G}-45.141)+1216.551}$ | $4.6920\times10^3$ | 0.90869 |
| GH4169 at 650℃ | $\ln(N_f)=0.302+\dfrac{11.625(\varepsilon_a\frac{\tau_a}{G}+297.975\frac{\tau_a}{G}+185.977)}{(\varepsilon_a+\gamma_a)(\varepsilon_a\frac{\tau_a}{G}+297.975\frac{\tau_a}{G}+185.977)+1202.181}$ | $2.9412\times10^2$ | 0.96022 |

Figure 4 shows the difference between the actual fatigue life cycles and the predicted fatigue life cycles derived from the optimal expressions in Table 4.

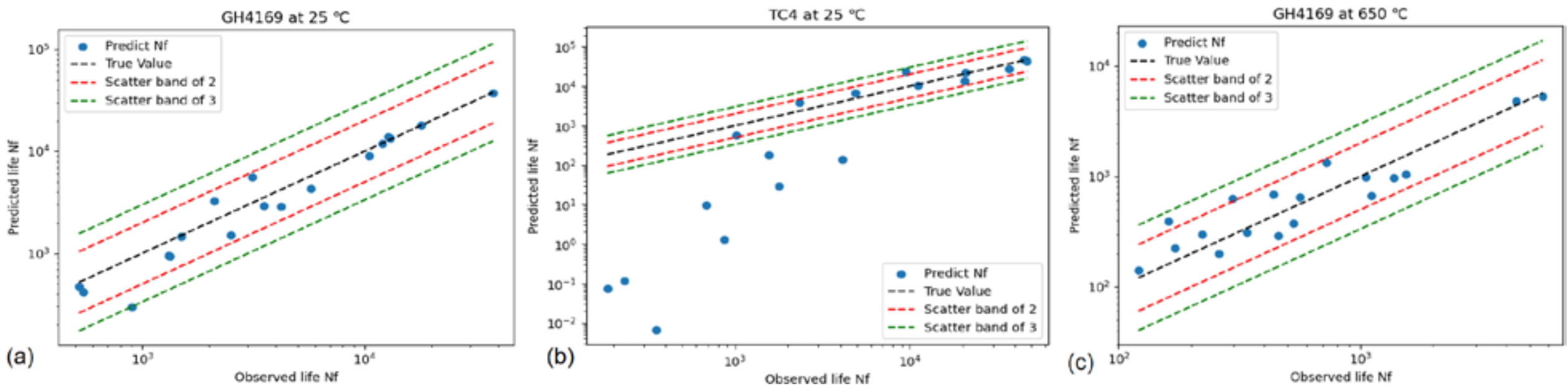

**Figure 4. Retaining the GH4169 optimal prediction expression structure as a generic structure to explore the optimal prediction expressions for both GH4169 and TC4 under different operating conditions. Comparisons of the observed experimental and predicted fatigue lives. (a) GH4169 estimated at 25℃, (b) TC4 estimated at 25℃, (c) GH4169 estimated at 650℃.**

Figure 4 illustrates that for GH4169 across varying temperatures, the majority of cyclic life predictions made by RSL fall within a twofold error margin of the actual cyclic life, with only a limited number of points residing within the two- to threefold error range. However, when this expression is applied to predict the fatigue life of TC4, the predictive performance notably diminishes due to the difference in material types. This observation indicates that the RSL approach is effective in capturing the relationship between fatigue life and mechanical performance parameters within the experimental dataset. For the same material, the fatigue life prediction formula identified by RSL demonstrates applicability across different operational conditions, where the impact of these conditions is primarily reflected in the adjustment of parameter values. Conversely, when applied to different materials, the prediction formula reveals fundamental structural differences. To resolve this dilemma, we need to explore new generalized life prediction formula for the corresponding material (e.g., TC4 in this study) according to fatigue experimental data of the material under different operating conditions.

To mitigate the risk of overfitting in the RSL algorithm, a Ten-Fold Cross-Validation was conducted on the material dataset to evaluate the model's generalization capability. The number of every validation dataset during the ten runs could be 1 or 2 (18*0.1), and the validation data from all the runs were aggregated and plotted against the predicted versus actual values. As illustrated in Fig. 5, the trends observed in the ten validation results are in alignment with the predictions depicted in Fig. 4. For GH4169 at 25°C, over 90% of the predicted values from the validation data fall within the twofold error band, with an RMSE of $9.196 \times 10^3$ across the ten validations. This indicates that RSL method has a robust generalization capability, effectively predicting unknown data and ensuring the reliability and robustness of the model.

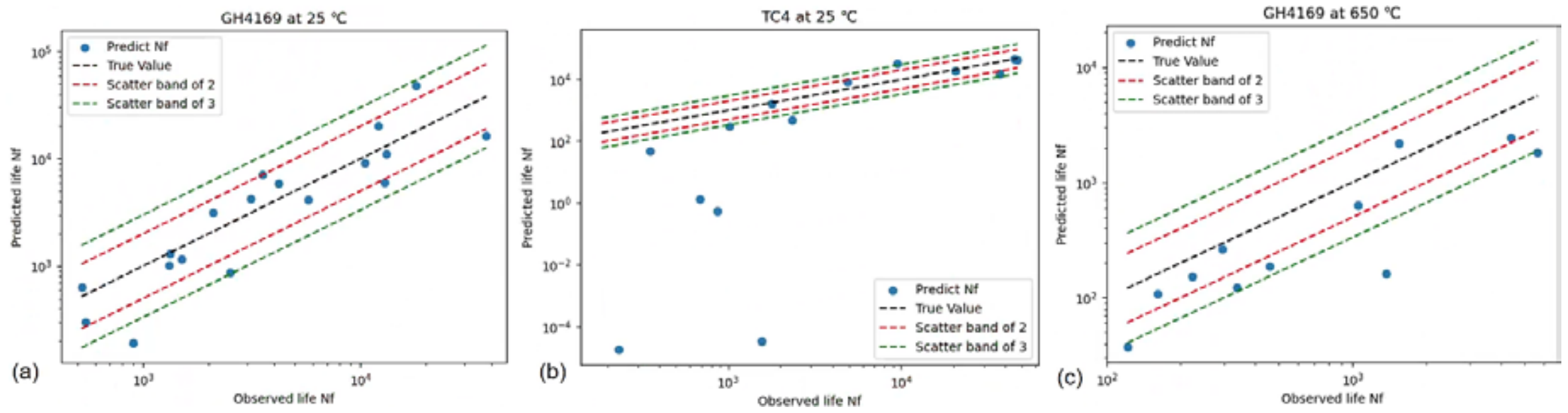

**Figure 5. Under the ten-fold cross-validation, the developed predictive formula developed using GH4169 is adopted to predict both GH4169 and TC4 under various operating conditions for (a) GH4169 estimated at 25℃, (b) TC4 estimated at 25℃, (c) GH4169 estimated at 650℃.**

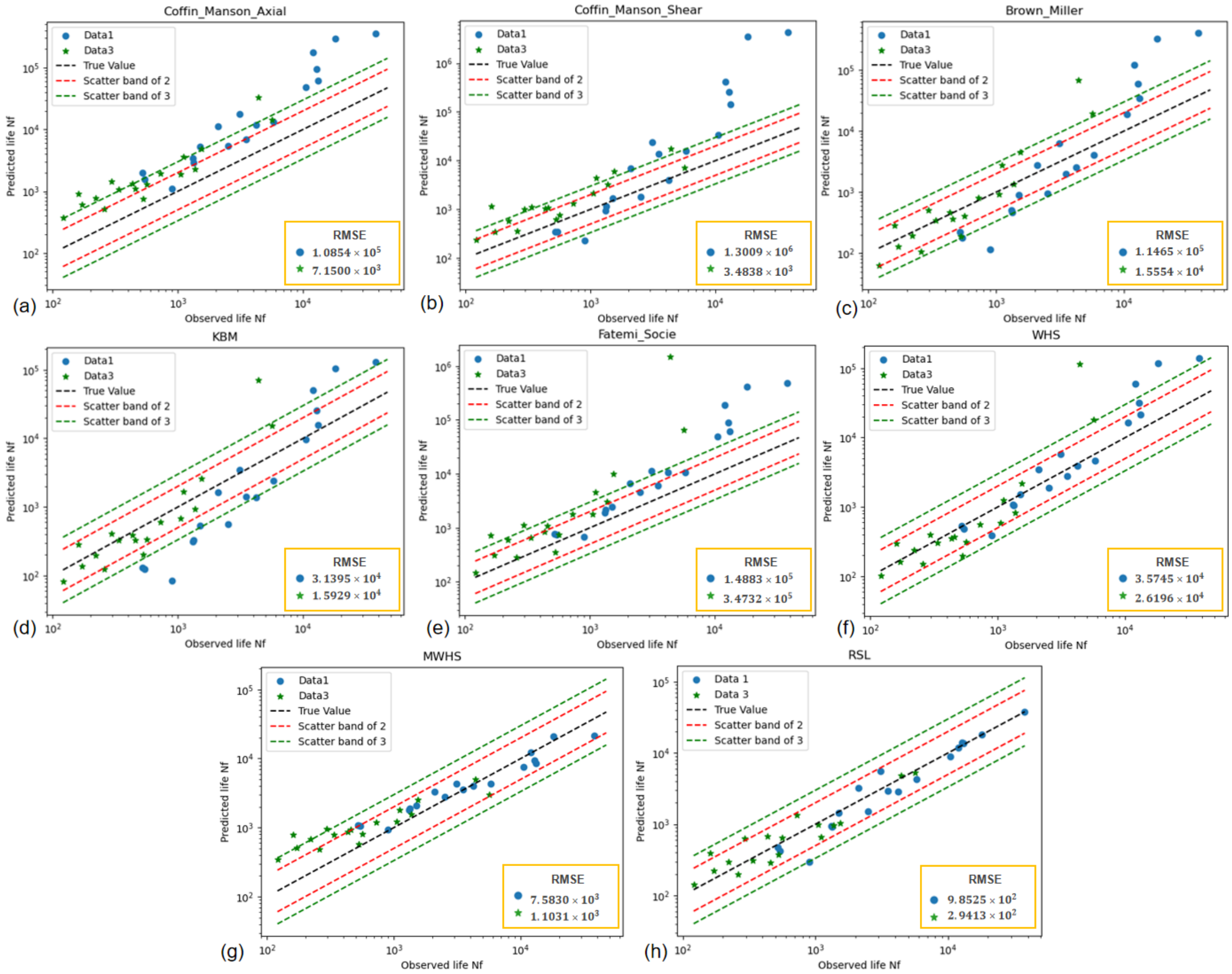

**Figure 6. Fatigue life prediction of GH4169 at 25°C (Data 1) and 650°C (Data 3) using six empirical formulas and the predictive formula derived by RSL. Comparisons between observed experimental and**

**predicted fatigue lives are shown from: (a) Coffin-Manson criterion (axial), (b) Coffin-Manson criterion (shear), (c) Brown-Miller criterion (BM), (d) Kandil-Brown-Miller criterion (KBM), (e) Fatemi-Socie criterion (FS), (f) WHS criterion, (g) MWHS criterion, and (h) RSL-derived expressions.**

**(3) Comparative experiments**

**Case A: Comparison with empirical methods**

A comparison was conducted between the predictive performance of the RSL method and the empirical formulas discussed in Section 2.1. This was achieved by applying the datasets Data1 and Data3 for GH4169 at 25°C and 650°C, respectively, to predict the cyclic life. Figure 6 presents a comprehensive comparison between the predicted life values derived from diverse methodologies and the observed experimental outcomes. The error band ranges are determined by the maximum value and minimum value of the observed life in all the datasets, plotting their double and triple ranges.

It was observed that the theoretical empirical formulas do not provide stable fatigue life predictions for specific types of materials in specific operating conditions, with a significant portion of the predicted values exceeding three times the actual values. Methods such as KBM, WHS, MWHS, and our RSL approach all exhibit strong tracking performance relative to the actual fatigue test data. However, the cyclic life predictions made by RSL are closer to the actual values, showing greater convergence in the graphical representation. By evaluating RMSE and $R^2$, we found that the RSL method achieved higher predictive accuracy (with RMSE $9.8525 \times 10^2$ and $2.9413 \times 10^2$) compared to the empirical formulas, resulting in the most accurate predictions.

**Case B: Comparison with data-driven methods**

In addition to comparing the predictive performance of RSL with theoretical empirical methods in Case A, the prediction results were also compared with several typical data-driven ML methods, including LSTM, SVR, and RF. Every dataset was divided into training and validation sets as inputs to ML methods in an 8:2 ratio, and all the data was predicted with the trained ML models. Final prediction results are shown in Figure 7.

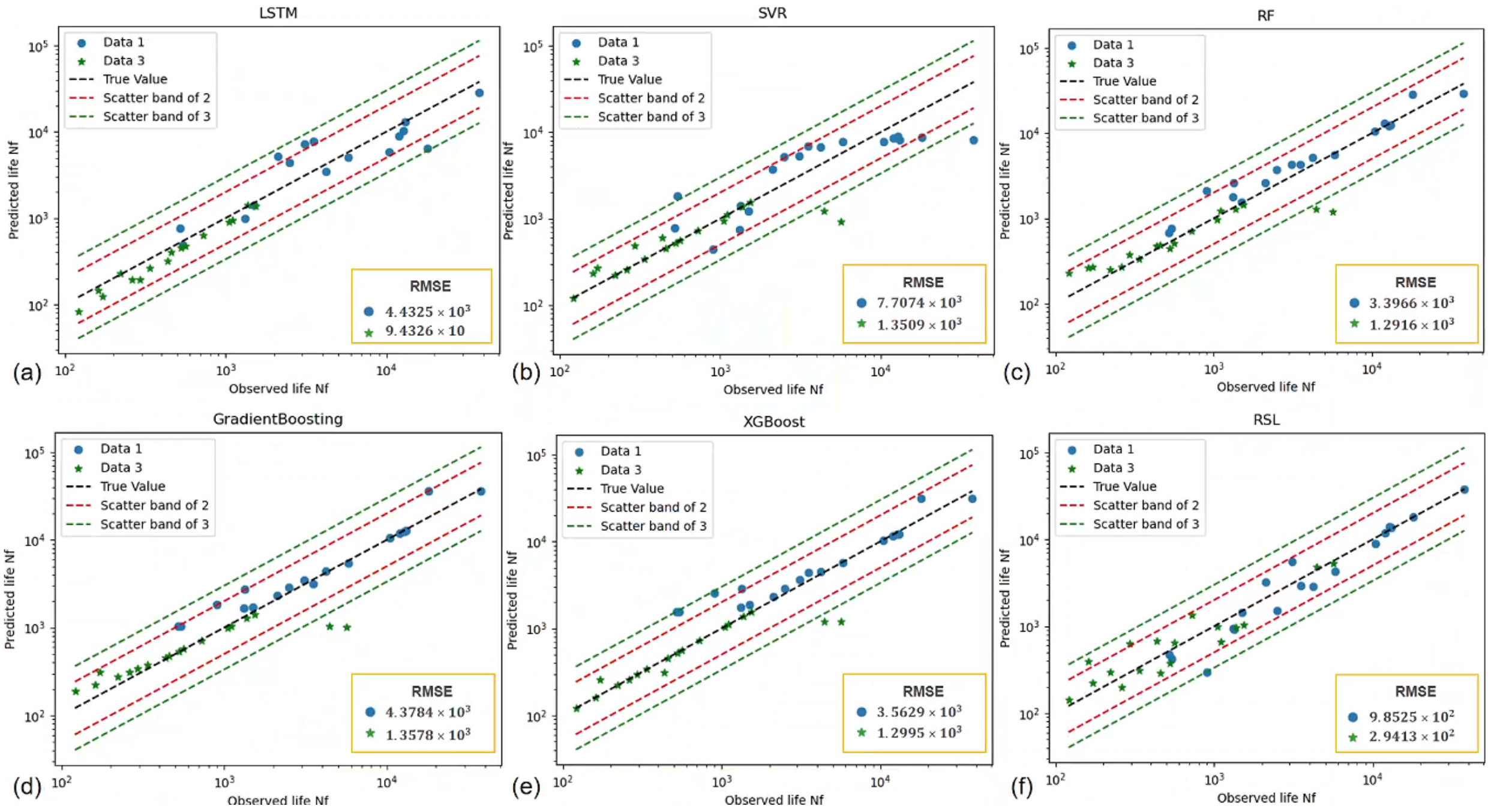

**Figure 7. Predicting the fatigue life of GH4169 at 25°C (Data 1) and 650°C (Data 3) using five machine learning models and the optimal prediction expressions explored by RSL. Comparisons of the observed experimental and predicted fatigue lives estimated by: (a) LSTM, (b) SVR, (c) RF, (d) Gradient Boosting, (e) XGBoost, (f) RSL expressions.**

As illustrated in the Fig. 7 above, the predictive accuracy of RSL for GH4169 fatigue life is comparable to, or even superior to, that of the majority of ML models. However, since ML algorithms essentially function as black boxes, they do not permit the explicit capture of the relationship between mechanical performance parameters and fatigue life. In light of the significance of interpretability, we prefer to use the SR algorithm to elucidate the underlying physical relationships between performance parameters.

## 4.3 Prediction of the turbine blade fatigue life

A fundamental aspect of estimating the lifespan of turbine blades is the prediction of fatigue life based on the mechanical properties of the materials used. Our RSL method has accurately identified the explicit physical relationship between the mechanical properties and fatigue life of turbine blade materials. By integrating these insights, once the mechanical performance parameters of the critical areas of the blade under various operating conditions have been determined, the fatigue life of the turbine blade can be accurately predicted.

(1) **Finite element simulation**

The turbine engine features a symmetrical annular structure, with each stage typically containing dozens to hundreds of blades, depending on the aspect ratio and other critical factors[33]. Each blade experiences similar loading conditions. To save computational time and costs, this project focuses on simulating a single blade using COMSOL, while simplifying certain structural elements that minimally impact the finite element simulation. The model of the single blade is shown in Fig. 8, with an overall radial length of 112.5 mm; the root of the blade is 21 mm high and 7.1 mm wide; and the blade crown measures 53.6 mm in length.

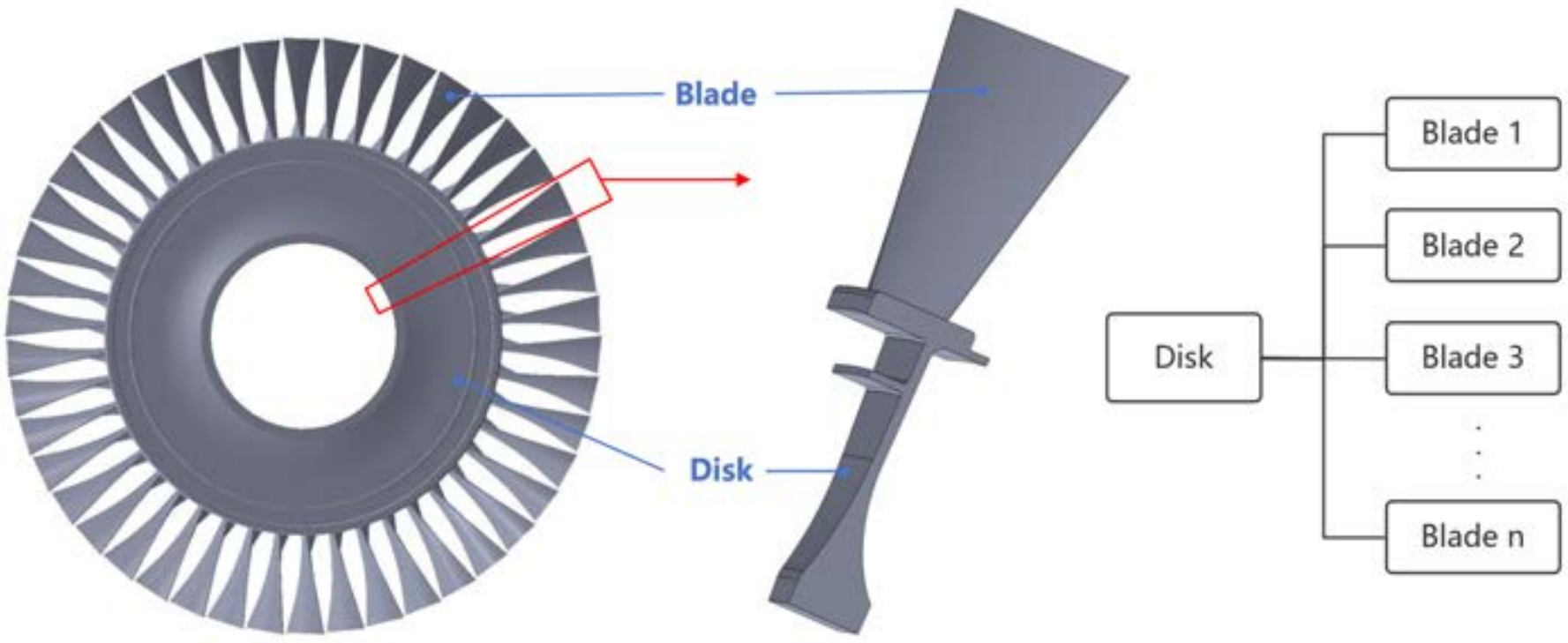

**Figure 8. High-pressure turbine blade structure.**

The commonly used casting material for turbine blades is the nickel-based superalloy GH4169. At the blade's maximum rotational speed, the temperature at the critical areas of the blade can reach approximately 600-700°C. Accordingly, the material properties of GH4169 at 650°C were employed as the simulation input for the blade model. Each blade is subjected to a multitude of complex stresses, including centrifugal, aerodynamic and thermal stresses, which must be simplified during the load simulation process[34]. Focusing on the fatigue life of the turbine blade under rotational speed conditions and its own weight, we simplified the loads on the blade model to centrifugal force and gravitational force. To simulate the actual operating conditions of the turbine blade, a symmetrically varying rotational speed was applied to the model, which results in a corresponding symmetric cyclic variation in centrifugal force. Furthermore, a gravitational force was applied to the blade, and the model was constrained by setting appropriate boundary conditions. The COMSOL simulation of the blade under these conditions is depicted in Fig. 9.

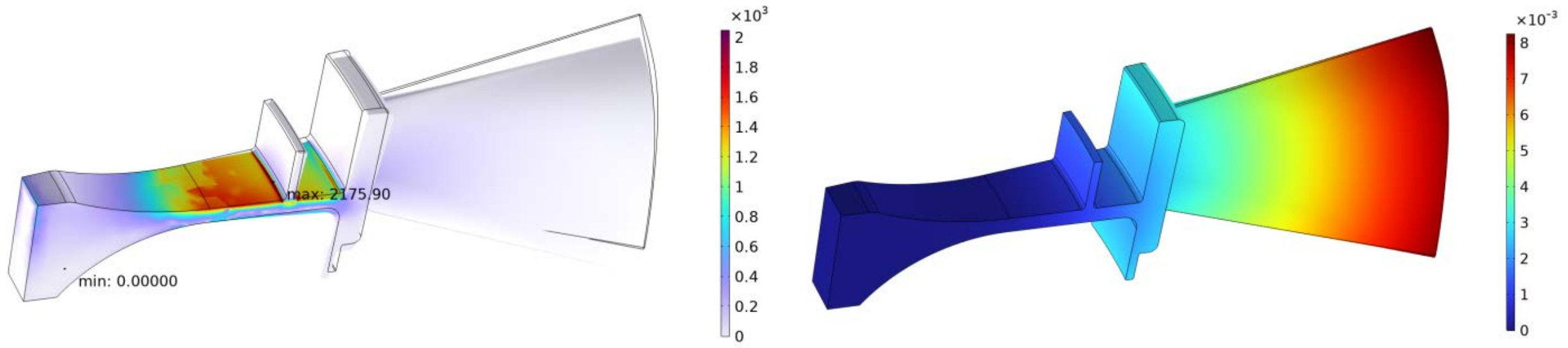

**Figure 9. Stress and strain results from the COMSOL simulation of the single turbine blade.**

By conducting simulations, we obtained the stress and strain conditions at various locations on the blade model. Using Von Mises stress as the evaluation criterion, we identified the most critical areas on the blade and determined the stress conditions at these high-risk locations.

(2) **Life prediction**

One of the primary causes of high-pressure turbine blade failure is low-cycle fatigue. During engine operation, under cyclic loading characterized by 'start-up—operation—shutdown', the surface temperature of the high-pressure turbine blade rapidly increases from room temperature to operational high temperatures, accompanied by a significant acceleration in the blade's rotational speed. This results in a rapid increase in centrifugal forces and other loads on the blade. Dynamic operating loads have the greatest impact on the creep fatigue life of turbine discs, which can lead to low-cycle fatigue[35], increasing the likelihood of crack initiation in stress concentration areas such as the blade root and blade crown. Considering the actual operating conditions of turbine engine blades, we conducted COMSOL simulations using the rotational speeds corresponding to four different operating conditions (S1, S2, S3, S4) as input variables for the blade model. The simulations identified the locations on the blade model that exhibited the highest Von Mises stress values, and the stress conditions at these critical locations were recorded and are presented in Table 5.

**Table 5.** Mechanical performance parameters at critical locations of the blade under different operating conditions.

| **Operating Conditions** | $\omega$(rpm) | $\varepsilon_{a,max}$(%) | $\gamma_{a,max}$(%) | $\sigma_{a,max}$(MPa) | $\tau_{a,max}$(MPa) |
|---|---|---|---|---|---|
| S1: Start-up -- Maximum Speed -- Start-up | 0-12000-0 | 1.76 | 0.62 | 2711.1 | 854.08 |
| S2: Start-up -- Cruise -- Start-up | 0-9500-0 | 1.10 | 0.39 | 1698.8 | 535.18 |
| S3: Idle -- Maximum Speed -- Idle | 8200-12000-8200 | 0.94 | 0.33 | 1445.6 | 455.42 |
| S4: Cruise -- Maximum Speed -- Cruise | 9500-12000-9500 | 0.66 | 0.23 | 1012.3 | 318.9 |

In this table, $\omega$ represents the actual rotational speed of the blade, while $\varepsilon_{a,max}$, $\gamma_{a,max}$, $\sigma_{a,max}$ and $\tau_{a,max}$ denotes the mechanical performance parameters at the critical locations of the blade (including the amplitude of axial strain, shear strain, axial stress, and shear stress).

In accordance with the fatigue life prediction comparison presented in Section 4.2, the optimal expression identified by RSL was selected, as it exhibited the most accurate prediction performance under symmetric cyclic loading. The aforementioned expression, identified as the optimal prediction formula for GH4169 at 650°C in Table 4, was employed to predict the fatigue life cycles by inputting the mechanical performance parameters of the critical blade locations into the formula. In order to ascertain the veracity of the RSL model's fatigue life predictions for the blade, as well as its predictive characteristics, further validation is required. The MWHS empirical formula and LSTM model, which demonstrated the most accurate predictive performance in Section 4.2, were selected to predict the blade's fatigue life. The results were then compared with those obtained by the RSL method. The comparative results of the three methods are presented in Table 6.

**Table 6. Comparison of fatigue life prediction effectiveness for the blade under different operating conditions between RSL, MWHS and LSTM.**

| Operating Conditions | $\omega$(rpm) | $N_f$ of RSL (cycles) | $N_f$ of MWHS (cycles) | $N_f$ of LSTM (cycles) |
|---|---|---|---|---|
| S1 | 0-12000-0 | 107 | 233 | 2921 |
| S2 | 0-9500-0 | 569 | 827 | 4129 |
| S3 | 8200-12000-8200 | 946 | 1370 | 4836 |
| S4 | 9500-12000-9500 | 2043 | 5168 | 7660 |

The comparative results demonstrate that under all four operating conditions, the fatigue life trends predicted by the empirical formula represented by MWHS, and the ML model represented by LSTM, are consistent with those predicted by the RSL method. As the variability in operating conditions for turbine blades increases, their fatigue life significantly decreases. In the S4 condition "Cruise - Maximum Speed - Cruise", where the operating conditions are the most stable, the predicted fatigue life is notably extended. The MWHS and RSL methods both predict fatigue life on the same dimensional scale, with minimal discrepancy. However, the LSTM model predicts a significantly higher number of life cycles, likely due to the tendency of ML algorithms to overfit smaller datasets, resulting in more aggressive predictions.

Further analysis reveals that the life cycles predicted by the RSL method are more conservative, whereas the ML models represented by LSTM, tend to produce more aggressive estimates. This distinction is particularly important in the context of fatigue life prediction for critical components in aircraft engines. Conservative life estimates can lead to more frequent scheduled inspections of turbine blades, thereby reducing the risk of potential failures and ensuring the reliability and safety of flight operations. As the fatigue test dataset grows, the accuracy of the RSL method will be further validated, providing a reliable reference for the maintenance and repair of critical components in future engineering applications, offering scientific guidance for life management of critical components.

# 5 Conclusion and future work

This paper introduces Reinforced Symbolic Learning (RSL), a novel symbolic regression algorithm guided by deep reinforcement learning and constrained by logical validity, to effectively model the relationship between mechanical properties and fatigue performance of turbine blades. In RSL, a binary tree structure and sub-node tracking array are used to constrain the expression structure, and logical constraint rules are introduced. Specifically, the overall length of the expression is controlled, the partial dimensionless preprocessing and function nesting constraints ensure the reasonableness of the physical meaning of the expression, and the limitations on the operation and number of constants are employed to enhance the simplicity of the expression. In addition, the logical constraint rules effectively reduce the search space, enhancing both the computational efficiency and accuracy of the algorithm. While RNN is used as a reinforcement learning carrier to generate the selection probability of each symbol in the symbol library, which serves as a reinforcement learning selection strategy. Selective reinforcement is applied to elite expressions from the expression batch, guiding subsequent batches to generate superior structures.

Using the widely employed turbine blade materials GH4169 and TC4 as evaluations and case studies, we identified structurally consistent and highly predictive formulas for each material. The results show that over 90% of the GH4169 predictions fall within a twofold error margin, confirming the high predictive accuracy of the RSL method. Compared to traditional empirical formulas, RSL uncovers more precise and explicit expressions tailored to the materials. Additionally, RSL matches or exceeds the predictive accuracy of most machine learning models, while offering greater interpretability and facilitating integration with existing theoretical frameworks. By incorporating finite element simulations to evaluate the mechanical properties under various conditions, we were able to predict the fatigue life cycles of the blades.

This paper is an initial effort of RSL for effective prediction of turbine blade life. It holds promise for application to a wide range of life-limited parts beyond turbine blades. The future work could focus on optimizing the algorithm to automatically determine the most suitable expression length based on dataset complexity. Given the current dataset limitations, the RSL algorithm has primarily explored generalized fatigue life prediction formulas for the same materials under varying conditions. As the dataset expands regarding size and varieties, it's expected to extend the RSL to uncover generalized predictive formulas that account for environmental variations across different materials.

# Declaration of competing interest

The authors declare that they have no known competing financial interests or personal relationships that could have appeared to influence the work reported in this paper.

# Data availability

Data will be made available on request.

# Acknowledgements

The work was supported by the National Key R&D Program of China (2022YFE0196400), the National Natural Science Foundation of China (52305288), the Zhejiang Provincial Natural Science Foundation of China under Grant No. LDT23E05013E05.

# Appendix

## Appendix A. Fatigue test data set of GH4169 and TC4

Tubular specimens with outer diameters of $\phi$14mm and $\phi$17mm and a length of 32mm were employed for multiaxial fatigue testing[12, 27], as illustrated in Fig. A1. The experiments were strain-controlled, with all specimens maintaining identical dimensions under the same testing conditions.

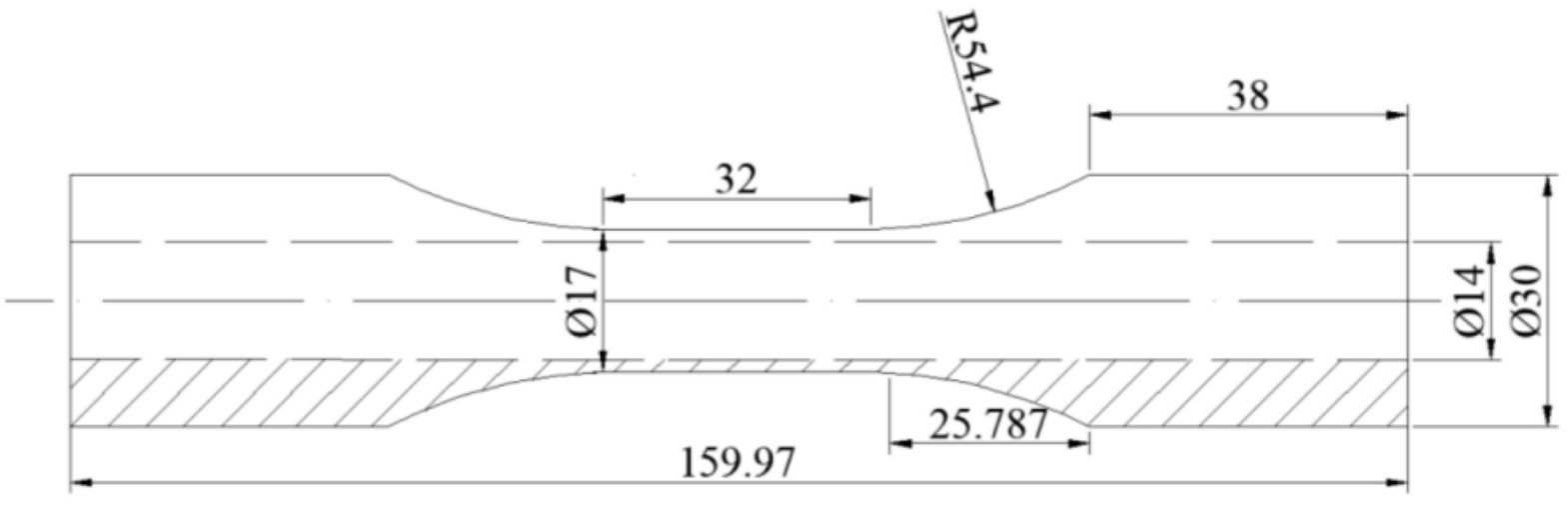

Figure A1. Fatigue test specimens under multiple axisymmetric load loading.

A fully reversed sinusoidal load with a frequency of 0.5-1.0 Hz was applied in axial, torsional, proportional, and 45°/90° non-proportional loading conditions. The experimental data for GH4169 were obtained under symmetric cyclic loading[27], while the data for TC4 were acquired under both symmetric and asymmetric cyclic loading conditions[12], as detailed in Table A4. The fatigue performance parameters of GH4169 and TC4 at room temperature and 650°C are presented in Tables A1, A2, and A3. Due to limitations in experimental data, certain shear fatigue properties were estimated using empirical formulas.

Table A1. Fatigue and cyclic stress–strain properties of Ni-based superalloy GH4169 at 25 $°C$.[27]

| | | | | | | |
|---|---|---|---|---|---|---|
| **Monotonic properties** | $E$/GPa | $G$/GPa | $\sigma_y$/MPa | $v_e$ | $K_1$/MPa | $n_1$ |
| | 198.5 | 67 | 1083.1 | 0.48 | 1579.7 | 0.06 |
| **Uniaxial properties** | $\sigma_f'$/MPa | $b$ | $\varepsilon_f'$ | $c$ | $K'$/MPa | $n'$ |
| | 1815.5 | -0.06 | 0.45 | -0.63 | 1892.3 | 0.078 |
| **Torsional properties** | $\tau_f'$/MPa | $b_1$ | $\gamma_f'$ | $c_1$ | $K_1'$ | $n_1'$ |
| | 1091.6 | -0.07 | 4.46 | -0.77 | 1047.1 | 0.099 |

Table A2. Fatigue and cyclic stress–strain properties of TC4 at 25 $°C$.[12]

| | | | | | | |
|---|---|---|---|---|---|---|
| **Monotonic properties** | $E$/GPa | $G$/GPa | $\sigma_y$/MPa | $v_e$ | $K_1$/MPa | $n_1$ |
| | 108.4 | 43.2 | 942.5 | 0.25 | 1054 | 0.0195 |
| **Uniaxial properties** | $\sigma_f'$/MPa | $b$ | $\varepsilon_f'$ | $c$ | $K'$/MPa | $n'$ |
| | 1116.9 | -0.049 | 0.579 | -0.679 | 1031 | 0.0478 |
| **Torsional properties** | $\tau_f'$/MPa | $b_1$ | $\gamma_f'$ | $c_1$ | $K_1'$ | $n_1'$ |
| | 716.9 | -0.06 | 2.24 | -0.8 | 446.7 | 0.016 |

Table A3. Fatigue and cyclic stress–strain properties of Ni-based superalloy GH4169 at 650 $°C$.[32]

| | | | | |
|---|---|---|---|---|
| **Monotonic properties** | $E$/GPa | $G$/GPa | $\sigma_y$/MPa | $v_e$ |
| | 182 | 62 | 1150 | 0.325 |
| **Uniaxial properties** | $\sigma_f'$/MPa | $b$ | $\varepsilon_f'$ | $c$ |
| | 1565.2 | -0.086 | 0.162 | -0.580 |
| **Torsional properties** | $\tau_f'$/MPa | $b_1$ | $\gamma_f'$ | $c_1$ |
| | 1091.6 | -0.06 | 1.62 | -0.75 |

Table A4. Fatigue data of GH4169 and TC4 at 25°C and 650°C (multi-axisymmetric load).[12, 27, 32]

| **Material** | **Temperature** | $\boldsymbol{\varphi}$**(° )** | $\boldsymbol{\varepsilon_a}$**(%)** | $\boldsymbol{\gamma_a}$**(%)** | $\boldsymbol{\sigma_a}$**(MPa)** | $\boldsymbol{\tau_a}$**(MPa)** | $\boldsymbol{N_f}$**(cycles)** |
|---|---|---|---|---|---|---|---|
| | | 0 | 1.221 | 1.598 | 937.7 | 478 | 901 |
| | | 0 | 0.770 | 1.175 | 857.1 | 463.3 | 1331 |
| | | 0 | 0.703 | 1.017 | 794.6 | 422.6 | 2503 |
| | | 0 | 0.612 | 0.880 | 788.6 | 404 | 4200 |
| | | 0 | 0.476 | 0.732 | 779.4 | 436.3 | 10456 |
| | | 0 | 0.342 | 0.622 | 666.9 | 418.6 | 18027 |
| | | 45 | 1.066 | 1.4151 | 1086.6 | 559 | 542 |
| | | 45 | 0.807 | 1.126 | 996.6 | 494 | 1315 |
| GH4169 | 25$°C$ | 45 | 0.521 | 0.965 | 796.3 | 519.2 | 3530 |
| | | 45 | 0.513 | 0.865 | 837.8 | 487.7 | 5764 |
| | | 45 | 0.423 | 0.714 | 790.3 | 454.9 | 13086 |
| | | 45 | 0.338 | 0.613 | 683.2 | 433.4 | 37904 |
| | | 90 | 1.069 | 1.3081 | 1199.5 | 659.3 | 520 |
| | | 90 | 0.712 | 1.0231 | 1062.5 | 602 | 1496 |
| | | 90 | 0.568 | 0.888 | 968.8 | 540.8 | 2102 |
| | | 90 | 0.492 | 0.827 | 951.1 | 554.7 | 3119 |
| | | 90 | 0.393 | 0.649 | 787.1 | 473.3 | 12008 |
| | | 90 | 0.406 | 0.686 | 822.1 | 502.8 | 12829 |
| TC4 | $25°C$ | 0 | 0.345 | 0.648 | 388.8 | 278.5 | 47195 |
| | | 0 | 0.427 | 0.710 | 466.4 | 296 | 20611 |
| | | 0 | 0.576 | 0.938 | 490.6 | 282.8 | 4141 |
| | | 0 | 0.687 | 1.111 | 532.1 | 312.7 | 1795 |
| | | 0 | 0.863 | 1.371 | 538.8 | 299.4 | 868 |
| | | 0 | 1.391 | 2.038 | 530.5 | 261 | 351 |
| | | 45 | 0.391 | 0.643 | 435.6 | 276.9 | 20953 |
| | | 45 | 0.418 | 0.702 | 472 | 303.2 | 9478 |
| | | 45 | 0.496 | 0.831 | 545.2 | 342.6 | 4898 |
| | | 45 | 0.620 | 1.043 | 592 | 340.9 | 1563 |
| | | 45 | 0.772 | 1.255 | 629 | 341.3 | 683 |
| | | 45 | 1.224 | 1.756 | 679.8 | 353.8 | 185 |
| | | 90 | 0.349 | 0.639 | 392.8 | 279.6 | 45138 |
| | | 90 | 0.418 | 0.704 | 475.7 | 307.8 | 37273 |
| | | 90 | 0.499 | 0.821 | 562.6 | 356.4 | 11152 |
| | | 90 | 0.556 | 0.934 | 623.6 | 401.2 | 2332 |
| | | 90 | 0.632 | 1.079 | 703.2 | 427.7 | 1017 |
| | | 90 | 1.229 | 1.700 | 678.6 | 382.3 | 233 |
| GH4169 | $650°C$ | 45 | 0.354 | 0.420 | 601 | 347 | 4420 |
| | | 90 | 0.397 | 0.479 | 679 | 434 | 5665 |
| | | 0 | 0.408 | 0.592 | 503 | 295 | 1544 |
| | | 45 | 0.524 | 0.745 | 658 | 560 | 722 |
| | | 45 | 0.553 | 0.813 | 691 | 436 | 295 |
| | | 90 | 0.548 | 0.833 | 762 | 475 | 436 |
| | | 90 | 0.586 | 0.838 | 801 | 506 | 563 |
| | | 0 | 0.546 | 0.884 | 584 | 301 | 458 |
| | | 45 | 0.704 | 1.090 | 793 | 477 | 171 |
| | | 45 | 0.701 | 1.160 | 757 | 492 | 260 |
| | | 90 | 0.783 | 1.330 | 899 | 607 | 121 |
| | | 0 | 0.540 | 0.896 | 745 | 317 | 338 |
| | | 0 | 0.536 | 0.945 | 642 | 401 | 161 |
| | | 0 | 0.427 | 0.633 | 637 | 268 | 1108 |
| | | 0 | 0.448 | 0.709 | 556 | 370 | 1370 |
| | | 45 | 0.478 | 0.749 | 655 | 426 | 1048 |
| | | 45 | 0.625 | 1.000 | 648 | 435 | 222 |
| | | 90 | 0.613 | 1.010 | 838 | 527 | 529 |

# Appendix B. Hyperparameters in ML models

We have selected five typical ML methods to compare with our RSL algorithm, and the ML methods and their corresponding hyperparameters are shown in Table B1.

Table B1. Hyperparameter settings in ML methods.

| **Models** | **Hyperparameter** | **Setting** | **Models** | **Hyperparameter** | **Setting** |
|---|---|---|---|---|---|
| LSTM | $n_1$ | 300 | RF | Number of Estimators | 100 |
| | Activation | ReLu | | Max Depth | None |
| | Dropout | 0.2 | | Min Samples Split | 2 |
| | $n_2$ | 300 | | Min Samples Leaf | 1 |
| | Optimizer | Adam | Gradient | Number of Estimators | 300 |
| | Learning Rate | 0.001 | Boosting | Max Depth | 3 |
| | Epochs | 2000 | | Learning Rate | 0.01 |
| | Batch Size | 1 | XGBoost | Number of Estimators | 300 |
| | Loss Function | MSE | | Max Depth | 5 |
| SVR | Kernel Function | RBF | | Learning Rate | 0.01 |
| | C | 1000 | | | |
| | Gamma | Scale | | | |